\documentclass[10pt,twocolumn,letterpaper]{article}

\usepackage[pagenumbers]{cvpr} 

\usepackage[pdftex]{graphicx}
\usepackage{amsmath}
\usepackage{amssymb}
\usepackage{booktabs}

\usepackage{colortbl}
\usepackage{epsfig}
\usepackage{kotex} 
\usepackage{adjustbox}
\usepackage{pifont}
\newcommand{\cmark}{\ding{51}}%
\newcommand{\xmark}{\ding{55}}%

\usepackage{xcolor}
\usepackage{algorithm}
\usepackage[noend]{algpseudocode}
\usepackage{listings}

\usepackage[accsupp]{axessibility}  

\usepackage[pagebackref,breaklinks,colorlinks]{hyperref}

\usepackage[capitalize]{cleveref}
\crefname{section}{Sec.}{Secs.}
\Crefname{section}{Section}{Sections}
\Crefname{table}{Table}{Tables}
\crefname{table}{Tab.}{Tabs.}

\def\cvprPaperID{1167} 
\def\confName{CVPR}
\def\confYear{2022}

\begin{document}

\title{Beyond Semantic to Instance Segmentation: Weakly-Supervised Instance Segmentation via Semantic Knowledge Transfer and Self-Refinement}

\author{
Beomyoung Kim$^{1}$\hspace{1.5em}YoonJoon Yoo$^{1,2}$\hspace{1.5em}Chaeeun Rhee$^{3}$\hspace{1.5em}Junmo Kim$^{4}$\\ \\
{NAVER CLOVA$^1$\hspace{3em}NAVER AI Lab$^2$\hspace{3em}Inha University$^3$\hspace{3em}KAIST$^4$}\\
}

\definecolor{darkgray}{rgb}{0.66, 0.66, 0.66}
\newcommand\yj[1]{{\textcolor{magenta}{#1}}}
\newcommand\by[1]{{\textcolor{blue}{#1}}}
\newcommand\todo[1]{{\textcolor{red}{#1}}}
\newcommand\remove[1]{{\textcolor{darkgray}{#1}}}
\definecolor{Gray}{gray}{0.9}

\definecolor{codegreen}{rgb}{0,0.6,0}
\definecolor{codegray}{rgb}{0.5,0.5,0.5}
\definecolor{codepurple}{rgb}{0.58,0,0.82}
\definecolor{backcolour}{rgb}{0.97,0.97,0.94}

\lstdefinestyle{mystyle}{
    backgroundcolor=\color{backcolour},   
    commentstyle=\color{codegreen},
    keywordstyle=\color{magenta},
    numberstyle=\tiny\color{codegray},
    stringstyle=\color{codepurple},
    basicstyle=\ttfamily\tiny,
    breakatwhitespace=false,         
    breaklines=true,                 
    captionpos=b,                    
    keepspaces=true,                 
    numbers=left,                    
    numbersep=5pt,                  
    showspaces=false,                
    showstringspaces=false,
    showtabs=false,                  
    tabsize=2
}

\lstset{style=mystyle}

\maketitle

\begin{abstract}
    Weakly-supervised instance segmentation (WSIS) has been considered as a more challenging task than weakly-supervised semantic segmentation (WSSS). Compared to WSSS, WSIS requires instance-wise localization, which is difficult to extract from image-level labels.
    To tackle the problem, most WSIS approaches use off-the-shelf proposal techniques that require pre-training with instance or object level labels, deviating the fundamental definition of the fully-image-level supervised setting.
    In this paper, we propose a novel approach including two innovative components.
    First, we propose a \textit{semantic knowledge transfer} to obtain pseudo instance labels by transferring the knowledge of WSSS to WSIS while eliminating the need for the off-the-shelf proposals.
    Second, we propose a \textit{self-refinement} method to refine the pseudo instance labels in a self-supervised scheme and to use the refined labels for training in an online manner.
    Here, we discover an erroneous phenomenon, \textit{semantic drift}, that occurred by the missing instances in pseudo instance labels categorized as background class.  
    This \textit{semantic drift} occurs confusion between background and instance in training and consequently degrades the segmentation performance. We term this problem as \textit{semantic drift problem} and show that our proposed \textit{self-refinement} method eliminates the semantic drift problem.
    The extensive experiments on \textit{PASCAL VOC 2012} and \textit{MS COCO} demonstrate the effectiveness of our approach, and we achieve a considerable performance without off-the-shelf proposal techniques.
    The code is available at \url{https://github.com/clovaai/BESTIE}.
\end{abstract}

\section{Introduction}
\label{sec:intro}
The recent line of weakly-supervised semantic segmentation (WSSS) approaches~\cite{(drs)kim2021drs, (pmm)li2021pseudo, (eps)lee2021railroad, (rrm)zhang2020reliability} have achieved impressive performance enhancement, with commonly using class activation maps (CAMs)~\cite{(cam)zhou2016learning} to obtain class-wise localization maps from image-level labels.
However, weakly-supervised instance segmentation (WSIS) using image-level labels is still an open task because the CAM does not provide instance-wise localization maps.

\begin{figure}[t]
    \centering
    \includegraphics[width=\linewidth]{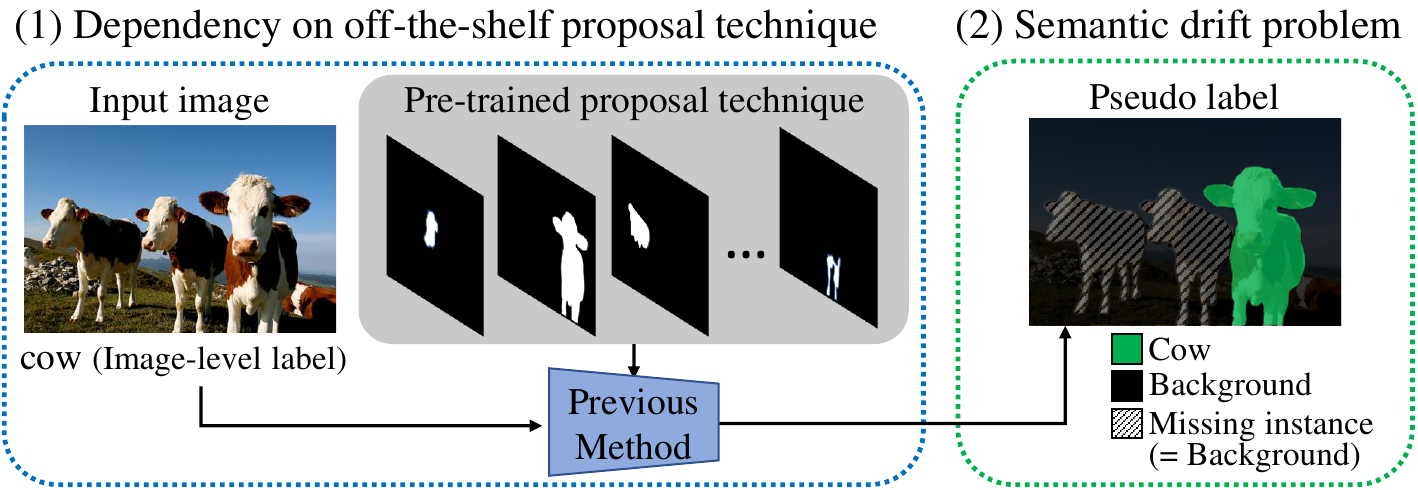}
    \caption{Two limitations of previous WSIS methods: (1) dependency on off-the-shelf proposal technique that requires pre-training with high-level labels including object or instance information. (2) semantic drift problem caused by the missing instances in the pseudo label guided as \textit{background} class. Our proposed BESTIE aims to solve the two problems simultaneously.
    }
    \label{fig:prev_limit}
    \vspace{-2mm}
\end{figure}

To extract the instance-wise information, most WSIS methods use off-the-shelf proposal techniques.
PRM~\cite{(prm)zhou2018weakly} takes suitable instance masks from segment proposals generated by MCG~\cite{(mcg)pont2016multiscale} and generates pseudo instance labels.
Also, LIID~\cite{(liid)liu2020leveraging} utilizes a pre-trained salient instance segmentor~\cite{(s4net)fan2019s4net} which produces class-agnostic instance-level masks.
We note that MCG and salient instance segmentor used in each method require pre-training with object boundary labels and class-agnostic instance masks, respectively.

However, the proposal-guided WSIS methods have two limitations as shown in Figure \ref{fig:prev_limit}.
First, their dependency on the off-the-shelf proposal techniques is considerably high, and it makes the methods difficult to apply to other specific domains such as medical images since the proposal techniques mostly target general objects. 
Furthermore, in a strict sense, the use of the proposal techniques trained by object or instance level information deviates from the definition of fully-image-level supervised segmentation.
Second, these methods cannot cope with the performance degradation caused by noisy pseudo-labels containing missing instances ($i.e.,$ false-negatives). As shown in Figure \ref{fig:prev_limit}, the left two missing cows are guided to the \textit{background} class and the right cow is guided to the \textit{cow} class, although all cows have semantically similar visual cues. We call this problem as \textit{semantic drift problem}.
This semantic drift between background and instance confuses the network and deteriorates the stable training convergence.

In this paper, we propose a new WSIS method, \textbf{BESTIE}: BEyond Semantic segmentation To InstancE Segmentation. BESTIE deprecates the use of off-the-shelf techniques to strictly follow a fully-image-level supervised setting. Also, BESTIE alleviates the semantic drift problem.
To solve the two problems, BESTIE proposes two innovative components, \textit{semantic knowledge transfer} and \textit{self-refinement}.

Specifically, in \textit{semantic knowledge transfer}, we transfer the knowledge of WSSS, which is relatively profoundly studied, to WSIS to generate the rough pseudo instance labels.
To obtain the instance cues from image-level labels, we propose  \textit{peak attention module (PAM)} that makes the CAM highlight the sparse representative region of objects. 
We note that the proposed components only use image-level labels, including the WSSS, and this eliminates the need for the off-the-shelf proposal technique.
Furthermore, to address the semantic drift problem, we introduce \textit{instance-aware guidance} that dynamically assigns the guidance region only to the labeled instance region.
This strategy allows more stable training of the network and progressively captures instance-level information of the missing instances.
Along with this strategy, to further refine the pseudo labels, we propose the self-supervised instance label refinement method that converts false-negatives in the pseudo labels to true-positives by a self-supervised manner and reflects them to the training in an online manner.
This method, shortly named \textit{self-refinement}, improves the quality of the pseudo labels as training progresses.

The extensive experiments on PASCAL VOC 2012~\cite{(voc2012)everingham2010pascal} and MS COCO 2017~\cite{(coco)lin2014microsoft} show the effectiveness of the proposed design. 
Even without the off-the-shelf proposal techniques, our method achieves a state-of-the-art performance of 51.0\% $mAP_{50}$ on VOC 2012 and 28.0\% $AP_{50}$ on COCO dataset.
Furthermore, we model the point-supervised instance segmentation by replacing the instance cues with point labels and can further boost the performance with an economical annotation cost.

Our contribution can be summarized as follows: 
\begin{itemize}
    \item We propose a novel WSIS method only using image-level labels, strictly following the fully-image-level supervised setting without the help of the proposal techniques pre-trained by object or instance level labels.
    \item We design the \textit{semantic knowledge transfer} strategy to obtain pseudo instance labels. This transfers the knowledge of WSSS and instance cues, extracted from the proposed PAM, to WSIS while eliminating the use of off-the-shelf proposal techniques.
    \item We propose a \textit{self-refinement} method to refine the pseudo instance labels in a self-supervised manner and reflect them back to the training in an online manner. Here, we introduce the \textit{instance-aware guidance} strategy to resolve the \textit{semantic drift} problem newly discovered in this paper.
\end{itemize}

\section{Related Work}

\subsection{Weakly-Supervised Semantic Segmentation}
Most weakly-supervised semantic segmentation (WSSS) studies handling image-level labels use CAMs~\cite{(cam)zhou2016learning} to localize class-wise object regions.
However, CAMs mainly focus on the sparse and discriminative object regions.
To address this issue, recent WSSS methods have proposed many approaches to expand activation regions.
AE-PSL~\cite{(psl)wei2017object} removes discriminative object regions by shifting attention to adjacent non-discriminative regions.
DRS~\cite{(drs)kim2021drs} proposes a module suppressing discriminative regions to expand activation regions.
However, these approaches are dependent on the off-the-shelf guidance, $i.e.$, saliency map.
To eliminate the use of the saliency map, saliency map-free methods has been also proposed:
RRM~\cite{(rrm)zhang2020reliability} proposes an end-to-end network jointly producing both CAMs and segmentation output to generate pseudo labels from only reliable pixels.
PMM~\cite{(pmm)li2021pseudo} proposes the proportional pseudo-mask generation by variation smoothing.
Also, some approaches have shown the expansion possibility of the WSSS to various domains such as medical~\cite{chan2019histosegnet} and satellite~\cite{nivaggioli2019weakly} images.

\subsection{Instance Segmentation}
Unlike semantic segmentation categorizing the pixel region in class-level, instance segmentation requires an instance-level mask.
The most widely used approach is a box-based two-stage method, $e.g.,$ Mask R-CNN~\cite{(mrcnn)he2017mask}, that predicts bounding boxes and then extracts the instance mask for each bounding box; this approach has reigned on the throne with state-of-the-art performance.
Recently, for the simple instance segmentation process, box-free one-stage instance segmentation methods~\cite{(inst_embedding)neven2019instance, (panopticdeeplab)cheng2020panoptic} have been proposed.
They represent each instance using 2-dimensional (2D) offset vectors; the pixels covering each instance are represented as 2D offset vectors directed to the center of each instance.
The center point of each instance is extracted from the center heatmap~\cite{(panopticdeeplab)cheng2020panoptic} or by clustering 2D offset vectors~\cite{(inst_embedding)neven2019instance} and the instance mask is obtained from instance grouping with the center point and 2D offset vectors.

\begin{figure*}[t]
    \centering
    \includegraphics[width=\linewidth]{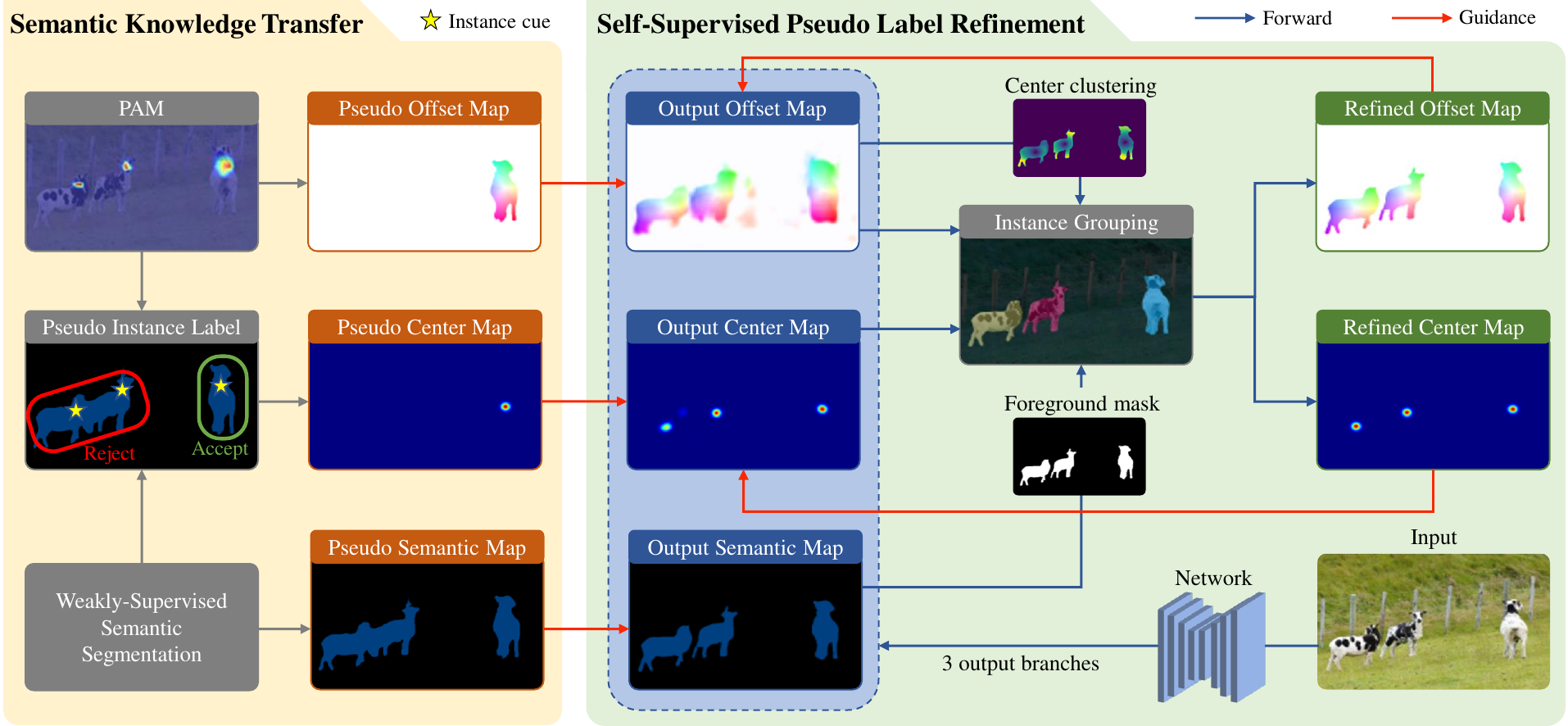}
    \caption{
    The overview of our framework including two-innovated steps: \textit{semantic knowledge transfer} and \textit{self-refinement}.
    In \textit{semantic knowledge transfer}, we obtain pseudo instance labels from the knowledge of WSSS and instance cues.
    Here, we obtain instance cues using \textit{peak attention module (PAM)}.
    In the \textit{self-refinement}, the network refines the pseudo labels in a self-supervised scheme and reflects them to the training in an online manner.
    For the stable training alleviating the semantic drift problem, we apply \textit{instance-aware guidance} strategy.
    We note that the entire process only uses the image-level labels.
    }
    \label{fig:overview}
    \vspace{-3mm}
\end{figure*}

\subsection{Weakly-Supervised Instance Segmentation}

The major difficulty in solving weakly-supervised instance segmentation (WSIS) occurs in the process of obtaining instance-level information from image-level labels.
To solve the problem, PRM~\cite{(prm)zhou2018weakly} produces a peak response map using the proposed peak back-propagation and then selects appropriate segment proposals generated by MCG~\cite{(mcg)pont2016multiscale}.
Arun $et$ $al.$~\cite{(consistent)arun2020weakly} defines the uncertainty in the pseudo label generation with the help of the segment proposals and iteratively train the network using the pseudo labels in an offline manner.
Fan $et$ $al.$~\cite{fan2018associating} and LIID~\cite{(liid)liu2020leveraging}) use the salient instance segmentor~\cite{(s4net)fan2019s4net} that produces class-agnostic instance-level masks when they generate pseudo labels.
In addition, box-based two-stage approaches \cite{(penet)ge2019label, (community)hwang2021weakly} use a selective search~\cite{(selective)uijlings2013selective} method to generate box proposals.
However, the off-the-shelf proposal techniques require pre-training with high-level supervision: class-agnostic object boundary for MCG and class-agnostic instance mask for the salient instance segmentor.
Also, since these proposal techniques target general objects, it disturbs their utilization in other domains such as a medical image.
IRN~\cite{(irn)ahn2019weakly} proposes a proposal-free method focusing on class-equivalence relations between a pair of pixels and represents instance-level information using their displacement field.
However, IRN has trouble in obtaining accurate instance-level information because the inter-pixel relations used in IRN are based on inter-class, not inter-instance.
To the best of our knowledge, existing methods have not considered the semantic drift problem caused by missing instances in pseudo labels, one fundamental challenge that should be addressed for WSIS.
In this paper, we discover and tackle the semantic drift problem explicitly for the first time and achieve an improved result in a fully-image-level supervised setting.

\section{Proposed Method}
\label{section:method}

\subsection{Overview}
\label{section:overview}

As shown in the left part of Figure \ref{fig:overview}, we first obtain pseudo instance labels using the knowledge of WSSS and instance cues, and this process is called \textit{semantic knowledge transfer}.
Here, we extract instance cues from image-level labels using the proposed \textit{peak attention module (PAM)} module.
Then, we apply the self-supervised pseudo label refinement abbreviated as \textit{self-refinement} strategy that refines the pseudo instance labels in a self-supervised scheme and reflects them to the training in an online manner as described in the right part of Figure \ref{fig:overview}.
In order to resolve the semantic drift problem and ensure stable training, we introduce the \textit{instance-aware guidance} strategy.
We note that our framework only uses the image-level labels as our guidance source including the WSSS part to deprecate off-the-shelf proposal techniques in the entire process.
We also provide the Pytorch-style pseudo-code for each proposed component in our supplementary material, showing that our method is quite easy to be implemented and simple.

\subsection{Preliminary: Instance Representation}
\label{section:preliminary}
Motivated by Panoptic-DeepLab \cite{(panopticdeeplab)cheng2020panoptic}, we represent an instance as a center point and corresponding 2D offset vectors. The 2D offset vectors direct the center point of each instance.
By adopting this representation method, we construct an instance segmentation network following the architecture of Panoptic-DeepLab, which consists of three output branches: semantic segmentation map, center map, and offset map.
The semantic segmentation map determines the foreground region.
For the post-processing, we extract center points of each instance from the center map; pixel locations with the same value before and after the max-pooling of the center map are regarded as center points.
Then, we allocate the ID of each instance in a pixel-level, and this procedure is called as instance grouping;
when the extracted $n$-th center point is denoted as $(x_{n}, y_{n})$ and the offset map is denoted as $\mathcal{O}(i, j)$ at the pixel location $(i,j)$, the instance ID $k_{i,j}$ at the pixel $(i, j)$ becomes
\begin{equation}
    \label{eq:grouping}
    k_{i,j} = \underset{k}{\mathrm{argmin}} ||(x_{k}, y_{k}) - ((i,j)+\mathcal{O}(i,j))||.
\end{equation}

\begin{figure}[t]
    \centering
    \includegraphics[width=\linewidth]{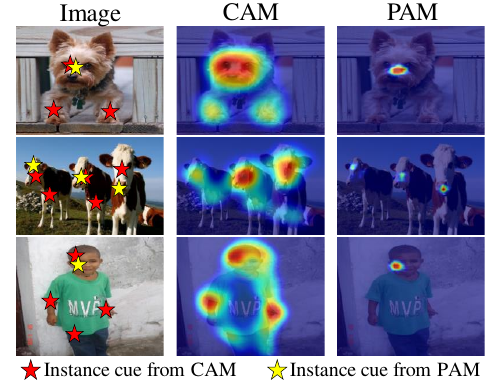}
    \caption{The activation maps from CAM and PAM. PAM helps extract more accurate instance cues than CAM.}
    \label{fig:cam_comparison}
    \vspace{-2mm}
\end{figure}

\subsection{Semantic Knowledge Transfer}
\label{section:semantic_knowledge_transfer}
The proposed \textit{semantic knowledge transfer} transfers the knowledge of WSSS to WSIS.
For the transfer, we rethink the two conditions of semantic and instance segmentation.
The \textbf{first condition} is that instance segmentation should separate overlapping instances of the same class, unlike semantic segmentation.
The \textbf{second condition} is that semantic segmentation and instance segmentation are equivalent to each other when instances of the same class do not overlap.
Based on these conditions, we generate pseudo instance labels utilizing the knowledge of WSSS.
From the WSSS output and instance presence cues, we check whether instances overlap.
Then, we select a non-overlapping instance mask as a pseudo instance mask, described in the left part of Figure \ref{fig:overview}.
Specifically, by performing the connected component labeling (CCL) algorithm~\cite{(ccl)he2009fast} on the WSSS output of each class, we obtain instance mask candidates and check how many instance cues are included in each instance mask candidate.
Following the \textbf{second condition}, the instance mask candidate with only one instance cue is selected as the pseudo instance mask.

\begin{figure}[t]
    \centering
        \includegraphics[width=\linewidth]{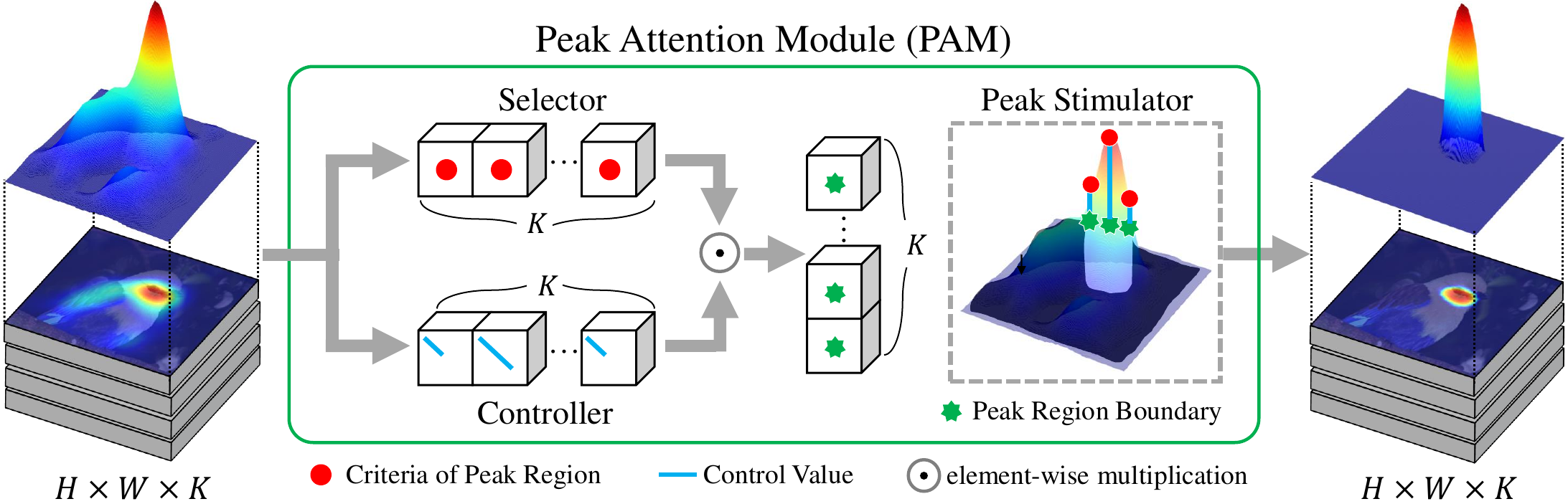}
    \caption{PAM architecture. From the selector, criteria of peak regions are selected. For a better explanation, three of them are illustrated in red points. Then, the controller determines how much to strengthen the attention on peak regions with control values, and each value is illustrated as the length of a blue line. Using criteria points and control values, the boundary of peak regions is set, and the stimulator strengthens the attention on peak regions by deactivating noisy regions whose values are lower than the boundary.}
    \label{fig:DRE}
\end{figure}

\begin{figure*}[t]
    \centering
    \includegraphics[width=\linewidth]{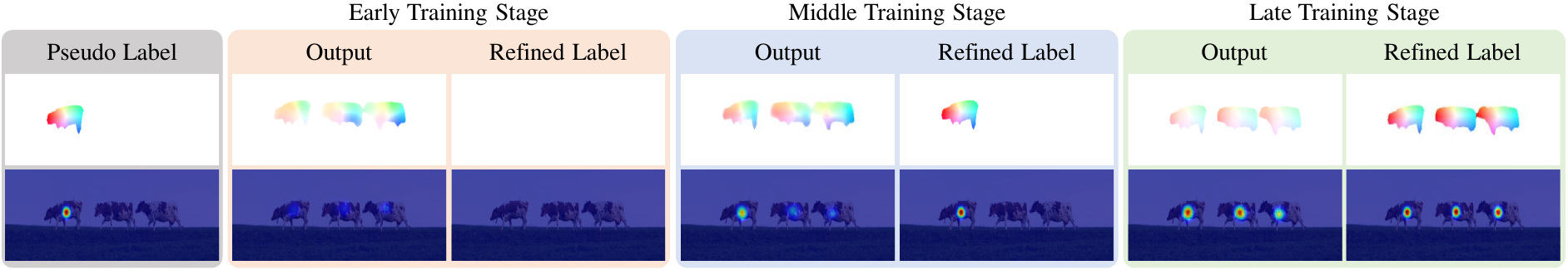}
    \caption{Comparison of offset and center maps. 
    As training iteration progresses, the network generates the higher-quality refined label.
    }
    \label{fig:label_comparison}
\end{figure*}

For the proper \textit{semantic knowledge transfer}, we require the accurate instance cue extraction method only using image-level labels.
The previous work, PRM~\cite{(prm)zhou2018weakly}, extracts the instance cue from CAMs~\cite{(cam)zhou2016learning}.
However, CAMs have a limitation in obtaining the accurate instance cue because several instance cues might be extracted in a single instance due to noisy activation regions as illustrated in Figure \ref{fig:cam_comparison}.
It disturbs the generation of pseudo instance labels since it violates the \textbf{second condition}.

To address this limitation, we propose a \textit{peak attention module (PAM)} to extract one appropriate instance cue per instance, motivated by DRS~\cite{(drs)kim2021drs}.
DRS suppresses discriminative object regions, spreading attention to adjacent non-discriminative regions in a self-supervised manner.
Contrary to the DRS, our PAM aims to strengthen the attention on peak regions, while weakening the attention on noisy activation regions.
PAM consists of three parts, as illustrated in Figure \ref{fig:DRE}: \textit{selector}, \textit{controller}, and peak \textit{stimulator}.
We denote the intermediate feature map as $X \in \mathbb{R}^{H \times W \times K}$, where $H$, $W$, and $K$ are the height, width, and the number of channels of $X$, respectively.
The \textit{selector} selects criteria points of peak regions using the global max pooling of $X$, and the criteria points are denoted as $S_{p} \in \mathbb{R}^{1 \times 1 \times K}$.
The \textit{controller} determines how much strengthen the attention on peak regions and is denoted as $G_{p} \in [0, 1]^{1 \times 1 \times K}$.
We strengthen the attention on peak regions by deactivating the attention on noisy regions.
In particular, $\tau_{p} = S_{p} \cdot G_{p}$ plays a role of the boundary of peak regions where $\cdot$ is an element-wise multiplication.
The regions in $X$ higher than $\tau_{p}$ are regarded as peak regions, otherwise regarded as noisy regions.
We deactivate the noisy regions by setting the value to zero, focusing on peak regions.
For the controller, we adopt the non-learnable setting of DRS, that is, all elements of $G_{p}$ are set to a constant value $\alpha$; the $\alpha$ is set to 0.7, and we find out that changes in the $\alpha$ between 0.3 and 0.7 do not significantly affect the performance of WSIS.
The PAM is plugged into the classifier, and we produce activation maps that localize the sparse representative region of each object as shown in Figure \ref{fig:cam_comparison} and then extract local maximum points ($i.e.,$ instance cues).
Note that our PAM does not require additional training parameters, and the classifier with PAM is optimized with a binary cross-entropy objective function by adaptively focusing on peak regions while increasing classification ability.

Combined with the knowledge of WSSS and instance cues extracted from PAM, we obtain pseudo instance masks and convert these masks into the pseudo center and offset maps following our instance representation method, as illustrated in Figure \ref{fig:overview}.
For the center map, the centroid point of each pseudo instance mask is encoded in a 2D Gaussian kernel with a standard deviation of 6 pixels.
For the offset map, all pixels in the pseudo instance mask contain 2D offset vectors directed to the corresponding center point.

\subsection{Instance-aware Guidance}
\label{section:instance_aware_guidance}
When training with the pseudo instance labels obtained by the \textit{semantic knowledge transfer}, we should handle the semantic drift problem.
Since the missing instances in the pseudo labels are guided as a \textit{background} class, semantic drift between background and instance deteriorates the stable training convergence.
To alleviate this problem, we introduce an \textit{instance-aware guidance} by taking the advantage of our instance representation method.

In our instance representation in Section~\ref{section:preliminary}, the offset and center maps represent instance-level information within the foreground region determined by the semantic segmentation map.
It means that the background region of the offset and center maps can be regarded as ignored regions.
Correspondingly, we dynamically assign the guidance region for the offset and center maps to only the region for the labeled instances;
this strategy is called the \textit{instance-aware guidance} and helps alleviate the semantic drift problem because the region of the offset and center maps for the missing instances ($e.g.,$ the white region of the pseudo offset map in Figure \ref{fig:overview}) is not reflected in the objective function.
Consequently, as shown in Figure \ref{fig:label_comparison}, the network can stably capture the instance-level information of the missing instances as the training progresses.

\subsection{Self-Supervised Pseudo Label Refinement}
\label{section:self-refinement}
Even we can alleviate the semantic drift problem, the number of true-positives in the pseudo labels is still not enough for training the network. For example, we can only have 30\% true-positives in the pseudo labels for VOC 2012.
Here, we propose a self-supervised pseudo label refinement strategy, abbreviated as \textit{self-refinement}, that refines the pseudo labels by converting false-negatives to true-positives in a self-supervised manner and reflects the refined labels to the training in an online manner.
The overall process of the \textit{self-refinement} is illustrated on the right side of Figure \ref{fig:overview}.
First, by training with the pseudo instance labels using \textit{instance-aware guidance} strategy, the network stably develops the generalization ability and gradually captures the instance-level information of the missing instances ($i.e.,$ false-negatives).
Next, following the Eq. (\ref{eq:grouping}), we perform the instance grouping using the network outputs.
Then, we generate refined offset and center maps from the instance mask created by instance grouping.
Last, the refined maps are used as guidance for the network.

For better refinement, we extract the center point by clustering the 2D offset vectors in the output offset map.
We call this process a \textit{center clustering} and explain the detailed algorithm in the supplementary material.
Even if the output center map misses some center points, we complement the refined center map using the clustered center points.

Using both pseudo labels and refined labels, we train the network.
We denote the network output semantic segmentation, offset map, and center map as $\mathcal{S}(\cdot)$, $\mathcal{O}(\cdot)$, and $\mathcal{C}(\cdot)$, respectively.
For the \textit{instance-aware guidance} of the offset and center maps, we collect sets of pixels for labeled instance regions from pseudo and refined labels, and each set is denoted as $\mathcal{P}_{pseudo}$ and $\mathcal{P}_{refined}$.
To utilize the refined labels as soft labels, we design a weight mask $\mathcal{W}(i,j)$:
\begin{equation}
\mathcal{W}^{n}(i,j) = \begin{cases} \mathcal{C}(x_{n}, y_{n}) & \text{$(i,j)$ $\in$ $\mathcal{P}^{n}_{pseudo}$,} \\ 0 & \text{otherwise,} \end{cases}
\end{equation}
where the center point of $n$-th instance in the refined labels are denoted as $(x_{n}, y_{n})$, and $\mathcal{C}(x_{n}, y_{n})$ means the confidence score of the $n$-th instance.
The $\mathcal{W}$ is used as the weight of the objective function for the refined labels.
The objective function of the center map is defined as:
\begin{equation}
  \begin{aligned}
    \mathcal{L}_{center} = \frac{1}{|\mathcal{P}_{pseudo}|} \sum_{(i,j) \in \mathcal{P}_{pseudo}}{(\mathcal{C}(i,j) - \hat{\mathcal{C}}(i,j))^2} +\\
    \frac{1}{|\mathcal{P}_{refined}|} \sum_{(i,j) \in \mathcal{P}_{refined}}{\mathcal{W}(i,j) \cdot (\mathcal{C}(i,j) - \bar{\mathcal{C}}(i,j))^2},
  \end{aligned}
\end{equation}
where the pseudo and refined center maps are $\hat{\mathcal{C}}(i,j)$ and $\bar{\mathcal{C}}(i,j)$, respectively.
Also, the objective function of the offset map is defined as:
\begin{equation}
  \begin{aligned}
    \mathcal{L}_{offset} = \frac{1}{|\mathcal{P}_{pseudo}|} \sum_{(i,j) \in \mathcal{P}_{pseudo}}{|\mathcal{O}(i,j) - \hat{\mathcal{O}}(i,j)|} +\\
    \frac{1}{|\mathcal{P}_{refined}|} \sum_{(i,j) \in \mathcal{P}_{refined}}{\mathcal{W}(i,j) \cdot |\mathcal{O}(i,j) - \bar{\mathcal{O}}(i,j)|},
  \end{aligned}
\end{equation}
where the pseudo and refined offset maps are $\hat{\mathcal{O}}(i,j)$ and $\bar{\mathcal{O}}(i,j)$, respectively.
Lastly, the objective function of the segmentation map is defined as:
\begin{equation}
  \begin{aligned}
    \mathcal{L}_{sem} = -\frac{1}{|\mathcal{P}_{sem}|} \sum_{(i,j) \in \mathcal{P}_{sem}}{log \mathcal{S}(i,j)},
  \end{aligned}
\end{equation}
where $\mathcal{S}$ is the output semantic map and $\mathcal{P}_{sem}$ is the set of all pixels in $\mathcal{S}$.
The network is jointly trained with the above objective functions, and the final objective function is:
\begin{equation}
  \begin{aligned}
    \mathcal{L} = \lambda_{center} \mathcal{L}_{center} + \lambda_{offset} \mathcal{L}_{offset} + \lambda_{sem} \mathcal{L}_{sem},
  \end{aligned}
\end{equation}
where $\lambda$ is a weight parameter, and set $\lambda_{center}=200$, $\lambda_{offset}=0.01$, and $\lambda_{sem}=1$ as used in \cite{(panopticdeeplab)cheng2020panoptic}.

Through our \textit{self-refinement} strategy, the pseudo labels can convert into high-quality refined labels.
The refined labels are generated from the network in an online manner at every mini-batch.
Also, since most of the processes are performed by GPU operation, the latency of the \textit{self-refinement} is negligibly small.


\begin{table}[t]
  \centering
  \caption{
    Quantitative comparison of state-of-the-art WSIS methods on VOC 2012 \textit{val}-set.
    $\dagger$ indicates applying MRCNN refinement.
    We denote the supervision sources as: $\mathcal{F}$ (full mask), $\mathcal{I}$ (image-level label), $\mathcal{P}$ (point), $\mathcal{C}$ (object count).
    The off-the-shelf proposal techniques are denoted as follows: $\mathcal{M}$ (segment proposal~\cite{(mcg)pont2016multiscale}), $\mathcal{R}$ (region proposal~\cite{(selective)uijlings2013selective}), $\mathcal{S_{I}}$ (salient instance segmentor~\cite{(s4net)fan2019s4net}).
  }
\label{tab:sota_instance}

  \begin{adjustbox}{max width=\linewidth}
  \begin{tabular}{c|c|c|cccc}
    \toprule
    Method & Sup & Extra & $mAP_{25}$ & $mAP_{50}$ & $mAP_{70}$ & $mAP_{75}$  \\
    \hline \hline
    Mask R-CNN~\cite{(mrcnn)he2017mask} & $\mathcal{F}$ & - & 76.7 & 67.9 & 52.5 & 44.9  \\
    \hline
    PRM~\cite{(prm)zhou2018weakly} &  $\mathcal{I}$ & $\mathcal{M}$ & 44.3 & 26.8 & - & 9.0   \\
    IAM~\cite{(iam)zhu2019learning} &  $\mathcal{I}$ & $\mathcal{M}$ & 45.9 & 28.3 & - & 11.9   \\
    Label-PEnet~\cite{(penet)ge2019label} &  $\mathcal{I}$ & $\mathcal{R}$ & 49.2 & 30.2 & - & 12.9   \\
    CL~\cite{(community)hwang2021weakly} &  $\mathcal{I}$ & $\mathcal{M}$, $\mathcal{R}$ & \textbf{56.6} & 38.1 & - & 12.3 \\
    \rowcolor{Gray} BESTIE (ours) & $\mathcal{I}$ & - & 53.5 & \textbf{41.8} & \textbf{28.3} & \textbf{24.2} \\
    \hline
    OCIS~\cite{(count)cholakkal2019object} &  $\mathcal{C}$ & $\mathcal{M}$ & 48.5 & 30.2 & - & 14.4 \\
    WISE-Net~\cite{laradji2020proposal} & $\mathcal{P}$ & $\mathcal{M}$ & 53.5 & 43.0 & -  & 25.9 \\
    \rowcolor{Gray} BESTIE (ours) & $\mathcal{P}$ & - & \textbf{58.6} & \textbf{46.7} & \textbf{33.1} & \textbf{26.3} \\
    \hline
    WISE$^{\dagger}$~\cite{(wise)laradji2019masks} & $\mathcal{I}$ & $\mathcal{M}$ & 49.2 & 41.7 & - & 23.7 \\
    IRN$^{\dagger}$~\cite{(irn)ahn2019weakly} & $\mathcal{I}$ & - & - & 46.7 & 23.5 & -  \\
    LIID$^{\dagger}$~\cite{(liid)liu2020leveraging} & $\mathcal{I}$ & $\mathcal{M}$, $\mathcal{S_{I}}$ & - & 48.4 & - & 24.9  \\
    Arun $et$ $al.$$^{\dagger}$~\cite{(consistent)arun2020weakly} & $\mathcal{I}$ & $\mathcal{M}$ & 59.7 & 50.9 & 30.2 & \textbf{28.5}  \\
    \rowcolor{Gray} BESTIE (ours)$^{\dagger}$ & $\mathcal{I}$ & - & \textbf{61.2} & \textbf{51.0} & \textbf{31.9} & 26.6 \\
    \rowcolor{Gray} BESTIE (ours)$^{\dagger}$ & $\mathcal{P}$ & - & \textbf{66.4} & \textbf{56.1} & \textbf{36.5} & \textbf{30.2} \\
    \bottomrule
  \end{tabular}
  \end{adjustbox}
\end{table}

\begin{table}[t]
  \centering
  \caption{
    Quantitative comparison of state-of-the-art WSIS methods on MS COCO 2017 dataset.
  }
\label{tab:sota_instance_coco}
  \begin{adjustbox}{max width=0.9\linewidth}
  \begin{tabular}{c|c|c|ccc}
    \toprule
    Method & Sup & Extra & $AP$ & $AP_{50}$ & $AP_{75}$  \\
    \hline \hline
    \multicolumn{6}{c}{\textit{\textbf{COCO val2017}}} \\
    Mask R-CNN~\cite{(mrcnn)he2017mask} & $\mathcal{F}$ & - & 35.4 & 57.3 & 37.5 \\
    WS-JDS~\cite{(ws-jds)shen2019cyclic} & $\mathcal{I}$ & $\mathcal{M}$ & 6.1 & 11.7 & 5.5 \\
    WISE-Net~\cite{laradji2020proposal} & $\mathcal{P}$ &  $\mathcal{M}$ & 7.8 & 18.2 & 8.8 \\
    \rowcolor{Gray} BESTIE (ours) & $\mathcal{I}$ & - & \textbf{14.3} & \textbf{28.0} & \textbf{13.2} \\
    \rowcolor{Gray} BESTIE (ours) & $\mathcal{P}$ & - & \textbf{17.7} & \textbf{34.0} & \textbf{16.4} \\
    \hline
    \multicolumn{6}{c}{\textit{\textbf{COCO test-dev}}} \\
    Mask R-CNN~\cite{(mrcnn)he2017mask} & $\mathcal{F}$ & - & 35.7 & 58.0 & 37.8 \\
    Fan $et$ $al.$~\cite{fan2018associating} & $\mathcal{I}$ & $\mathcal{S_{I}}$ & 13.7 & 25.5 & 13.5 \\
    LIID~\cite{(liid)liu2020leveraging} & $\mathcal{I}$ & $\mathcal{M}$, $\mathcal{S_{I}}$ & \textbf{16.0} & 27.1 & \textbf{16.5} \\
    \rowcolor{Gray} BESTIE (ours) & $\mathcal{I}$ & - & 14.4 & \textbf{28.0} & 13.5 \\
    \rowcolor{Gray} BESTIE (ours) & $\mathcal{P}$ & - & \textbf{17.8} & \textbf{34.1} & \textbf{16.7} \\
    \bottomrule
  \end{tabular}
  \end{adjustbox}
  \vspace{-2mm}
\end{table}

\section{Experiments}

\subsection{Dataset and Evaluation Metrics}
    
We demonstrate the effectiveness of the proposed approach on Pascal VOC 2012~\cite{(voc2012)everingham2010pascal} and COCO~\cite{(coco)lin2014microsoft} datasets.
For VOC 2012 dataset, following the common practice in previous works~\cite{(irn)ahn2019weakly, (liid)liu2020leveraging}, we use the augmented dataset that contains 10,582 training and 1,449 validation images with 20 object categories.
COCO dataset consists of 115K training, 5K validation, and 20K testing images with 80 object categories.
We evaluate the performance using the mean average precision (mAP) with intersection-over-union (IoU) thresholds of 0.25, 0.5, 0.7, and 0.75 for VOC 2012 and averaged AP over IoU thresholds from 0.5 to 0.95 for COCO.

\begin{table*}[t]
  \begin{minipage}[b]{0.34\linewidth}
    \centering
    \caption{
        Effect of the proposed methods: PAM, IAG (\textit{instance-aware guidance}), refine (\textit{self-refinement}), and cluster (\textit{center clustering}).
    }
    \label{tab:ablation_analysis}
    \begin{adjustbox}{max width=\linewidth}
      \begin{tabular}{cccc|cccc}
        \toprule
        PAM & IAG & refine & cluster & $mAP_{50}$  \\
        \hline
        \xmark & \xmark & \xmark & \xmark & 12.9 \\
        \cmark & \xmark & \xmark & \xmark & 29.3 \\
        \cmark & \cmark & \xmark & \xmark & 39.2 \\
        \cmark & \cmark & \cmark & \xmark & 41.5 \\ 
        \cmark & \cmark & \cmark & \cmark & 41.8 \\
        \bottomrule
      \end{tabular}
    \end{adjustbox}
    \vspace{-1mm}
  \end{minipage}
  \hfill
  \begin{minipage}[b]{0.41\linewidth}
    \centering
    \caption{
        Analysis of the effect of WSSS result on our WSIS performance.
        We measure the $mAP_{50}$ instance segmentation performance according to WSSS methods.
    }
    \label{tab:ablation_WSSS}
    \centering
    \begin{adjustbox}{max width=0.98\linewidth}
      \begin{tabular}{c|c|c}
        \toprule
        \multicolumn{2}{c|}{Semantic Segmentation} & Instance Segmentation \\
        \hline
         WSSS method & mIoU & $mAP_{50}$ \\
        \hline
         SingleStage~\cite{(singlestage)araslanov2020single}  &  62.7  &  39.7  \\
         RRM~\cite{(rrm)zhang2020reliability} &  66.3  &  41.1  \\     
         PMM~\cite{(pmm)li2021pseudo}           &  70.0  &  41.8  \\
        \hline
        ground truth  & - & 49.4  \\
        \bottomrule
      \end{tabular}
    \end{adjustbox}
    \vspace{-1mm}
  \end{minipage}
  \hfill
  \begin{minipage}[b]{0.20\linewidth}
    \centering
    \caption{
        Effect of iterative training strategy on our method. 
        \textit{iters}=0 means without the iterative training strategy.
    }
    \label{tab:iterative_training}
    \begin{adjustbox}{max width=0.8\linewidth}
      \begin{tabular}{c|c}
        \toprule
        \# iters & $mAP_{50}$  \\
        \hline
        0 & 41.8 \\
        1 & 41.9 \\
        2 & 41.9 \\
        \bottomrule
      \end{tabular}
    \end{adjustbox}
    \vspace{-1mm}
  \end{minipage}
\end{table*}

\begin{table*}[t]
  \begin{minipage}[b]{0.54\linewidth}
    \centering
    \includegraphics[width=0.83\linewidth]{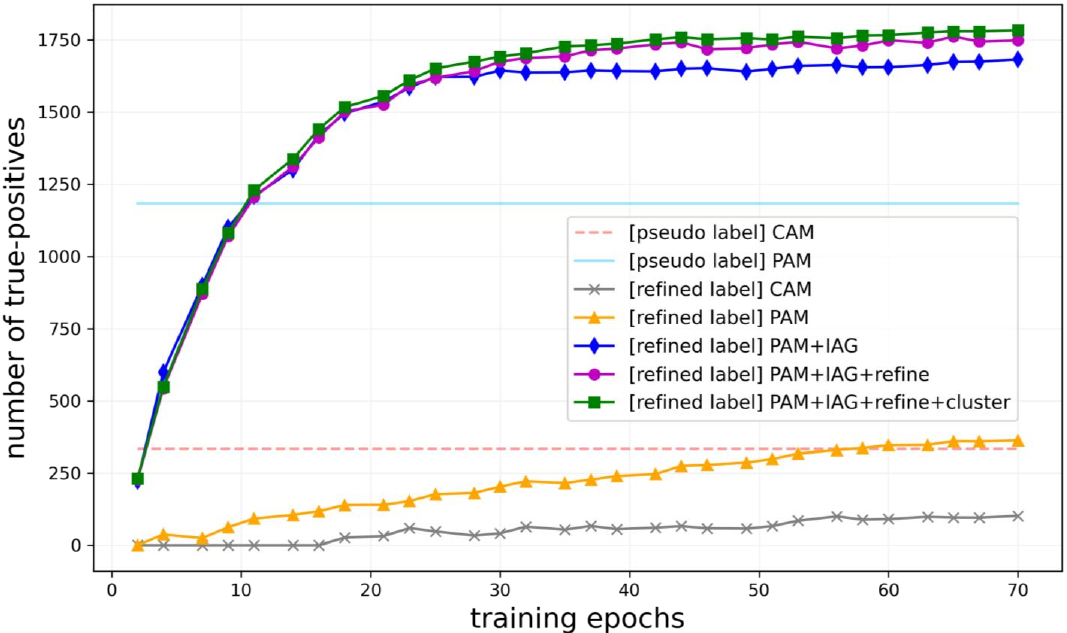}
    \captionof{figure}{Evolution of the number of true-positives on VOC 2012 \textit{train}-set.}
    \vspace{-2mm}
    \label{fig:evolution_tp}
  \end{minipage}
  \hfill
  \begin{minipage}[b]{0.45\linewidth}
    \centering
    \includegraphics[width=\linewidth]{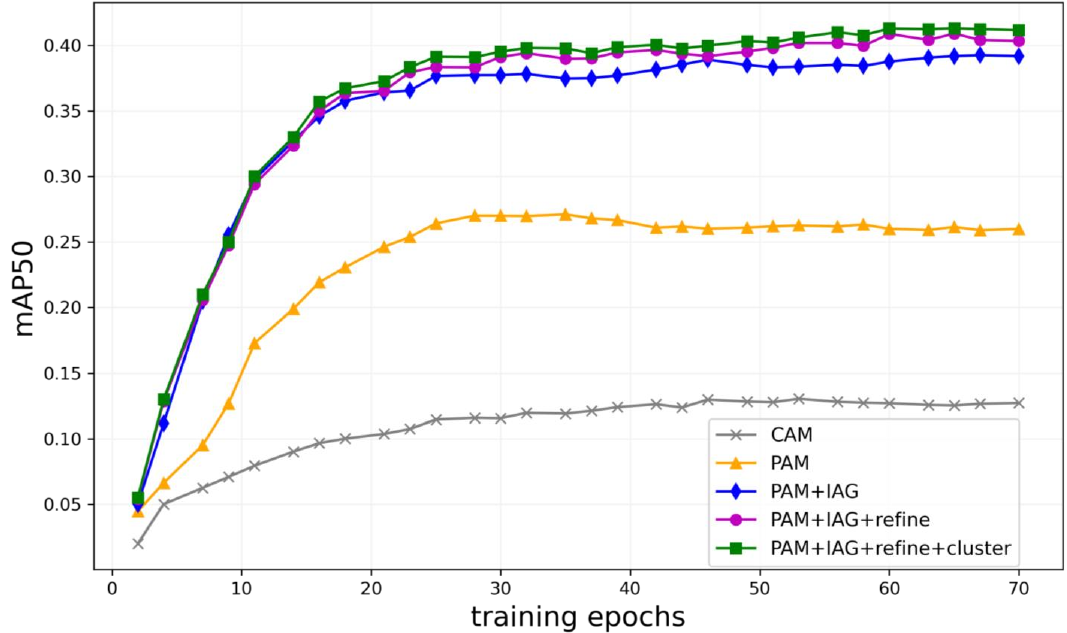}
    \captionof{figure}{Evolution of the $mAP_{50}$ on VOC 2012 \textit{val}-set.}
    \vspace{-2mm}
    \label{fig:evolution_map}
  \end{minipage}
\end{table*}

\subsection{Implementation Details}

For the \textit{semantic knowledge transfer}, we extract instance cues from the classifier equipped with our PAM.
We describe the detailed architecture of the classifier and analysis in the supplementary material.
For the fully-image-level supervised setting, we adopt PMM~\cite{(pmm)li2021pseudo} as our WSSS method because it does not use the saliency maps.

For the instance segmentation network, we follow the network structure of Panoptic-DeepLab~\cite{(panopticdeeplab)cheng2020panoptic} with a modification. 
We change the center map from class-agnostic to class-wise for more accurate instance grouping.
We adopt HRNet48~\cite{(hrnet)sun2019deep} as our backbone network.
The input size for training is 416${\times}$416, and we keep the original resolution for evaluation.
We train the network for 70 epochs with 32 batch size using Adam optimizer~\cite{(adam)kingma2014adam} with $5{\times}10^{-5}$ learning rate and polynomial learning rate scheduling~\cite{(polynomial)liu2015parsenet}.
Some approaches~\cite{(irn)ahn2019weakly, (liid)liu2020leveraging} employ an additional training step on Mask R-CNN~\cite{(mrcnn)he2017mask}; we denote this step as MRCNN refinement and train the Mask R-CNN following the official training recipe using pseudo labels generated by our network.
We used PyTorch 1.7 framework~\cite{(pytorch)paszke2019pytorch} with CUDA 10.1, CuDNN 7 and 8 NVIDIA V100 GPUs.

\subsection{Point-Supervised Instance Segmentation}
In our framework, point label can be used as weak supervision.
According to \cite{(pointsup)bearman2016s, (budget)bellver2019budget}, annotation costs are as follow: image-level (20.0 sec/img), object count (22.2 sec/img), point (23.3 sec/img), bounding box (38.1 sec/img), full mask (239.7 sec/img).
The point label is an economical label that is 16\% more expensive than the image-level label.
For our point-supervised setting, we replace instance cues from PAM with point labels for the \textit{semantic knowledge transfer} and replace the pseudo and refined center maps with the ground-truth center map for the \textit{self-refinement}.
Using the point supervision, we can obtain 10\% more true-positives in pseudo labels for VOC 2012 due to the accurate instance cue, boosting the performance as in Table \ref{tab:sota_instance}.

\subsection{State-of-the-arts Comparison}

\begin{figure*}[t]
    \centering
    \includegraphics[width=\linewidth]{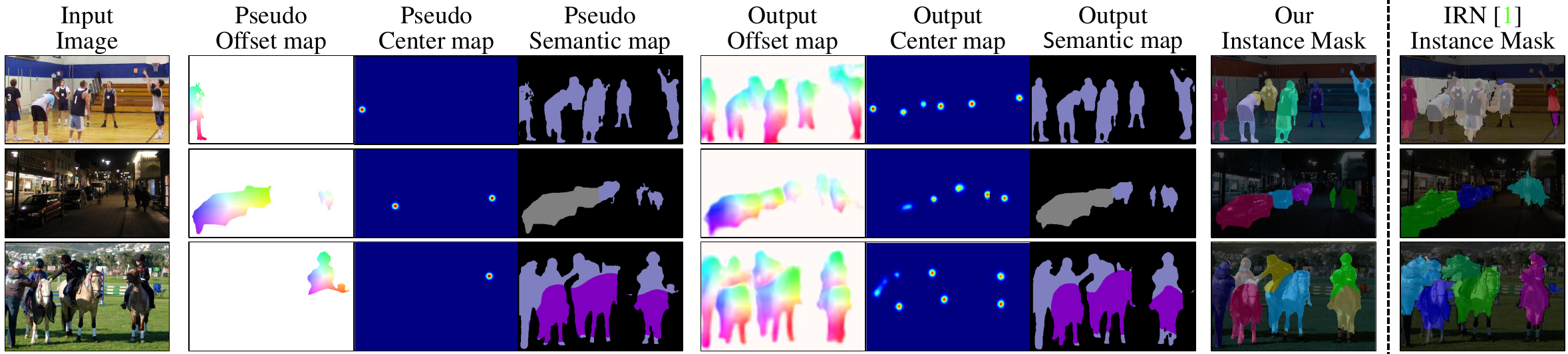}
    \vspace{-6mm}
    \caption{Qualitative results of our BESTIE trained with image-level supervision on VOC 2012 dataset.}
    \label{fig:qualitative_result}
    \vspace{-3mm}
\end{figure*}

We compare our BESTIE with existing state-of-the-art WSIS methods in Table \ref{tab:sota_instance} for VOC 2012 and Table \ref{tab:sota_instance_coco} for COCO dataset.
Even without the off-the-shelf proposals, BESTIE outperforms existing methods, especially in the $AP_{50}$ metric.
Although LIID~\cite{(liid)liu2020leveraging} achieved a 1.6\% $AP$ higher than ours on COCO, they utilized two proposal techniques that require pre-training with high-level labels, violating the fully-image-level supervised setting. 
Compared to the fully-image-level supervised method, IRN~\cite{(irn)ahn2019weakly}, we outperform the method (51.0\% $v.s.$ 46.7\%) because their displacement field, which is similar to our offset map, does not consider the semantic drift problem.
Also, IRN often fails to segment overlapping instances as in Figure \ref{fig:qualitative_result} because their inter-pixel relations are derived from class-wise information, not instance-wise information.
Given point supervision, we further increased the performance gap with other methods at a reasonable cost and achieve a new state-of-the-art performance on VOC and COCO datasets compared to the previous best point-supervised method, Wise-Net~\cite{(wise)laradji2019masks}.

\subsection{Ablation Study and Analysis}

For analysis, we skip the Mask R-CNN refinement and follow the mentioned implementation details.
We count the number of true-positives on VOC 2012 train set containing 1,464 images and 3,507 instances in Figure \ref{fig:evolution_tp} and measure the $mAP_{50}$ on VOC 2012 validation set in Figure \ref{fig:evolution_map}.

\vspace{1mm}\noindent{\textbf{Effect of PAM:}}
As shown in Figure \ref{fig:cam_comparison}, we have trouble in obtaining accurate instance cues from conventional CAM due to the noisy activation regions.
However, with our PAM, we can extract appropriate instance cues, which tremendously helps the proper \textit{semantic knowledge transfer} and obtain three times more true-positive training samples than the CAM as shown in Figure \ref{fig:evolution_tp}, giving a 16.4\% improvement as in the first and second rows of Table \ref{tab:ablation_analysis}.

\vspace{1mm}\noindent{\textbf{Effect of Instance-aware Guidance:}}
In this section, we abbreviate \textit{instance-aware guidance} as IAG.
For analysis, we train the network without IAG; it means that the whole region (including the background region) of the offset and center maps are reflected in the objective function.
As in the first and third rows of Table \ref{tab:ablation_analysis}, without IAG, the performance drops by 9.9\% because it suffers from the semantic drift problem as mentioned in the method section.
Also, as shown in Figure \ref{fig:evolution_map}, the model without IAG seems to be stuck in a local minimum, whereas the model with IAG seems to avoid the local minimum, enhancing the performance as training progresses.
This result convinces us that IAG is effective to alleviate the semantic drift problem.

\vspace{1mm}\noindent{\textbf{Effect of Self-Refinement:}}
Here, we compare the result of the network trained using only the pseudo labels without the \textit{self-refinement}.
As shown in Figure \ref{fig:evolution_tp}, the network with the \textit{self-refinement} can have more true-positives as training progresses.
Since the refined labels from the \textit{self-refinement} are guided to the training, the network can further capture instance-level features and improve the performance by 2.3\% as in the third and fourth rows of Table \ref{tab:ablation_analysis}.

\vspace{1mm}\noindent{\textbf{Effect of Center Clustering:}}
As in the fourth and last rows of Table \ref{tab:ablation_analysis}, a 0.3\% improvement when using the center clustering demonstrates that the center clustering can complement the generation of the refined labels.

\vspace{1mm}\noindent{\textbf{Influence of WSSS method:}}
The results in Table \ref{tab:ablation_WSSS} shows how the WSSS result affects the WSIS.
Originally, we adopt PMM~\cite{(pmm)li2021pseudo} for our WSSS method, which shows 70.0\% mIoU on VOC 2012 validation set.
Adopting the WSSS methods of 7.3\% and 3.7\% lower mIoU (SingleStage~\cite{(singlestage)araslanov2020single} and RRM~\cite{(rrm)zhang2020reliability}) drops $mAP_{50}$ by 2.1\% and 0.7\%.
The results show that the performance of WSIS is relatively robust to that of WSSS.
Additionally, we train with ground-truth semantic segmentation labels and obtain a performance gain of 7.6\% $mAP_{50}$; this result leaves us the opportunity that the advancement of the WSSS method can improve the performance of our approach.

\vspace{1mm}\noindent{\textbf{Does iterative training help?}}
Some weakly-supervised methods~\cite{(consistent)arun2020weakly, wang2018weakly, (simple)khoreva2017simple} maximize their performance by the iterative training strategy; they generate pseudo labels when training is finished and re-train the network using the pseudo labels in an offline manner.
This strategy gives a progressive improvement but requires a huge training complexity.
However, this strategy does not give us a noticeable improvement as in Table \ref{tab:iterative_training}, and we show that our single-step online \textit{self-refinement} is quite efficient for label refinement.

\vspace{1mm}\noindent{\textbf{Qualitative Results:}}
We provide some qualitative results in Figure \ref{fig:qualitative_result}.
Although the pseudo labels contain a few instance labels, BESTIE can accurately represent instance-level information, achieving high-quality instance masks.

\vspace{1mm}\noindent{\textbf{Limitation and Future direction:}}
Despite the significant performance enhancement from the proposed BESTIE, it remains more room to be improved.
In our method, the number of true-positives in the pseudo labels is limited by overlapping objects in the image (see Figure \ref{fig:qualitative_result}).
The number of overlapping objects is different from dataset to dataset, $i.e.,$ less for VOC dataset but many for COCO dataset, and this affects the performance in some sort. 
Although our method achieved promising results in the VOC dataset with small 30\% of true-positives in the pseudo labels, one future direction will be the suggestion of more effective true-positive acquiring rules for the various condition of data.

\section{Conclusion}

In this paper, we proposed a novel approach for WSIS by addressing the pain-points of previous methods: dependency on the off-the-shelf proposals where extra instance or object level guidance is indispensable and the semantic drift problem.
In our \textit{semantic knowledge transfer}, we transferred the knowledge of WSSS combined with instance cues to WSIS and obtained pseudo instance labels. 
Here, we proposed the PAM module to accurately extract the instance cues.
In our \textit{self-refinement}, we refined the pseudo labels in a self-supervised scheme and employed them in training.
For the stable learning with resolving the semantic drift problem, we introduced the \textit{instance-aware guidance} strategy.
Extensive experiments demonstrated the effectiveness of our approach, and our approach outperforms the previous methods with only using image-level labels and without any off-the-shelf proposals.
Lastly, we conclude that this research does not contain potential negative societal impact.

\section{Acknowledgement}
The authors thank NAVER Smart Machine Learning (NSML)~\cite{kim2018nsml} team for the GPU support.
This work was partly supported by Institute of Information \& communications Technology Planning \& Evaluation (IITP) grant funded by the Korea government(MSIT) (No.2021-0-02068, Artificial Intelligence Innovation Hub) and Basic Science Research Program through the National Research Foundation of Korea (NRF) funded by the Ministry of Science, ICT \& Future Planning (NRF-2021R1A2C2008946).

\nocite{(boxsup)dai2015boxsup}
\nocite{(scribble)lin2016scribblesup}
\nocite{(crawl)hong2017weakly}
\nocite{oh2017exploiting}
\nocite{choe2020attention}

{\small
\bibliographystyle{ieee_fullname}
\bibliography{ms}
}

\clearpage

\section*{Appendix: Detail of Center Clustering Algorithm}

\begin{figure}[h]
    \centering
    \includegraphics[width=\linewidth]{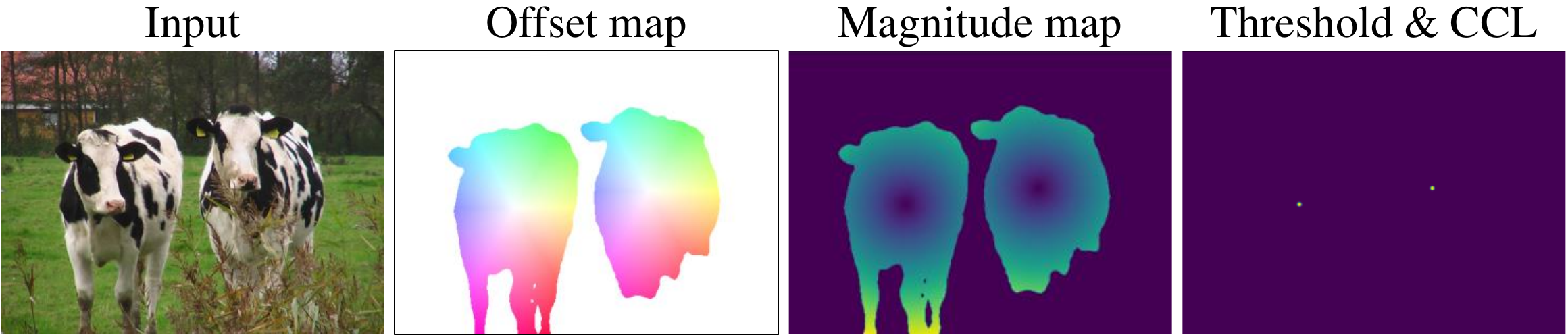}
    \caption{
    Illustration of the center clustering algorithm. 
    The blue and yellow pixels in the magnitude map indicate that pixel values are close to zero and far from zero, respectively.
    }
    \label{fig:clustering}
\end{figure}

For the complementary knowledge between each network output, we employ the center clustering algorithm to extract center points from the offset map when generating the refined label.
Here, we describe a detailed algorithm for the center clustering with a Figure \ref{fig:clustering}.
First, from the offset map, we create a magnitude map where each pixel represents the magnitude of the 2D vector.
In this magnitude map, the pixel near the center of each instance is close to zero.
Second, we apply a threshold to the magnitude map. We set the threshold to 2.5.
Last, we extract the center point of each mask candidate obtained from the connected component labeling (CCL) algorithm.   
Here, we observe that the optimal area of the mask candidate is determined according to the threshold.
For example, when the threshold is 2.5, the desired area of the mask candidate is near 21.
For reliability-check utilizing the above observation, we additionally check whether the area of the mask candidate is between 21-$\varepsilon$ and 21+$\varepsilon$; we empirically set the $\varepsilon$ to 3.
Due to the reliability-check process, we can prevent extracting false center points from the unstable offset map in the early training stage.

\section*{Appendix: Additional Ablation Study}

Here, we provide some additional ablation studies.
First is the class-wise center map. As mentioned in the experiment section, we modify the original Panoptic-DeepLab network; we change the class-agnostic center map to the class-wise center map.
This modification yields a 1.0\% $mAP_{50}$ improvement due to more accurate instance grouping; compared to the full supervision, our noisy offset vectors in the offset map are sometimes grouped with incorrect center points. To prevent incorrect instance grouping, we restrict the centers of other classes not to be grouped by adopting the class-wise center map.

The second is an additional analysis for the proposed methods. In Table 3 in the main paper, we provided the analysis for the proposed methods. Here, we show an additional study for the model without IAG but with self-refine.
As in Table \ref{tab:analysis_additional}, the self-refine without IAG drops the performance because the model without IAG suffers from the semantic drift problem. And the drift iteratively degrades the quality of the refined label, hurting the model.

The third is the effect of hyperparameter $\alpha$ that is a threshold for the PAM module.
When the $\alpha$ becomes large, more noisy regions are deactivated.
However, due to the IAG and self-refine, $mAP_{50}$ result of BESTIE is robust to the $\alpha$ as in Table \ref{tab:analysis_alpha}.

The fourth is the effect of the backbone network in BESTIE.
As mentioned in the experiment section, we adopt HRNet-48~\cite{(hrnet)sun2019deep} as our backbone network.
Here, we study the effect of the backbone network by replacing another backbone network, $i.e.,$ ResNet-50~\cite{(resnet)he2016deep}.
As an experimental result, the HRNet-48 backbone yields about 1\% $mAP_{50}$ higher performance than the ResNet-50 (41.8\% $mAP_{50}$ for HRNet-48 and 40.9\% $mAP_{50}$ ResNet-50 on VOC 2012 validation set).
The reason is that the receptive field of the HRNet-48 is much larger than the ResNet-50 and the HRNet-48 is a well-designed network for the key-point representation.

The last is a threshold for extracting instance cues from PAM.
Namely, we extract instance cues by obtaining local maximum points from the PAM.
Here, we adopt the instance cues whose value is larger than the threshold, which is set to 0.5.
When we change the threshold to 0.3 and 0.7, the number of true-positives in pseudo labels changes slightly, but the $mAP_{50}$ variation is quite small to $\pm$0.1\%.
This is because our \textit{self-refinement} method can progressively refine the pseudo labels and increase the number of true-positives.

\begin{table}[t]
    \centering
    \begin{minipage}{0.50\linewidth}
        \centering
        \caption{Additional analysis for the proposed methods.}
        \begin{adjustbox}{max width=\linewidth}
          \begin{tabular}{ccc|c}
            \toprule
            PAM & IAG & refine & $mAP_{50}$  \\
            \hline
            \checkmark &  &  &  29.3 \\
            \checkmark & \checkmark &   & 39.2 \\
            \checkmark & \checkmark & \checkmark & 41.8 \\
            \checkmark &  & \checkmark & 27.8 \\
            \bottomrule
          \end{tabular}
        \end{adjustbox}
        \label{tab:analysis_additional}
    \end{minipage}
    \hspace{5mm}
    \begin{minipage}{0.35\linewidth}
            \centering
            \caption{Effect of the hyperparameter $\alpha$.}
            \begin{adjustbox}{max width=\linewidth}
              \begin{tabular}{c|c}
                \toprule
                $\alpha$ & $mAP_{50}$ \\
                \hline
                0.3 & 41.76 \\
                0.5 & 41.80 \\
                0.7 & 41.72 \\
                \bottomrule
              \end{tabular}
            \end{adjustbox}
            \label{tab:analysis_alpha}
    \end{minipage}
\end{table}

\begin{figure*}[t]
    \centering
    \includegraphics[width=\linewidth]{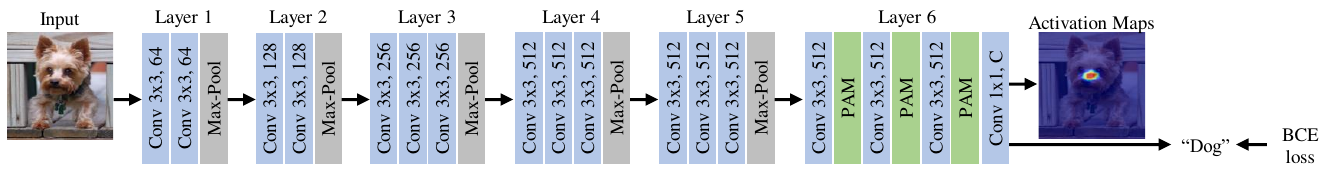}
    \caption{The detailed architecture of our classification network. BCE loss means the binary cross entropy loss function. The PAM is equipped in last three convolutional layers and trained with a self-supervised scheme.}
    \label{fig:classifier}
\end{figure*}

\section*{Appendix: Details of Peak Attention Module (PAM)}

\subsection*{Implementation Details}
As described in the main paper, we extract instance cues from the classifier with our PAM.
Here, we explain the implementation details of the PAM.
We employ VGG-16~\cite{(vgg)simonyan2014very} classifier and plugin our PAM into the last three convolutional layers of the classifier.
The architecture of the classifier with PAM is illustrated in Figure \ref{fig:classifier}.
For training the classifier with PAM, we use the binary cross-entropy loss function and the stochastic gradient descent (SGD) optimizer with a weight decay of 0.0005 and a momentum of 0.9.
The initial learning rate is set to 0.001 and is decreased by a factor of 10 at epoch 5 and 10. 
For data augmentation, images are randomly cropped to 321${\times}$321, and random horizontal flipping and random color jittering are applied.
We use a batch size of 5 and train the classifier for 15 epochs.
In the following section, we analyze how the PAM module affects each layer.

\subsection*{Effect of PAM on each layer of Classifier}
In this section, we analyze the effect of the PAM on each layer of the classifier.
With the classifier described in Figure \ref{fig:classifier} as our baseline, we plug-in or plug-out the module.
For the quantitative comparison as in Table \ref{tab:pam_module}, we evaluate the mean average precision (mAP) of our instance segmentation network without the Mask R-CNN refinement step.
Since our PAM strengthens the attention on peak regions by deactivating noisy regions, we necessary to accurately distinguish between peak and noise regions.
In lower-level layers ($i.e.,$ layer-1, layer-2, and layer-3), the classifier captures the local features such as edges, and the definition of the peak region is unclear, so the effect of the PAM is minor.
In contrast, in higher-level layers ($i.e.,$ layer-4, layer-5, and layer-6), especially in the last layer, the classifier captures the global features, and the distinction between peak regions and noisy regions is more clear; our PAM plays a meaningful role in the last layer.
From the results in Table \ref{tab:pam_module}, we note that the PAM equipped in only the last layer yields the best performance (41.8\% $mAP_{50}$) but the PAM equipped in the last three layers significantly degrades the performance (34.8\% $mAP_{50}$).
We conclude that it is most effective to use the PAM only in the last layer where the definition of peak regions and noisy regions is the most obvious, and excessive modulization of PAM might deactivate the important features, degrading the performance.

 \begin{table}[t]
    \centering
    \caption{
        Effect of the PAM on each layer of the classifier. $\checkmark$ means the PAM is equipped.
      }
    \begin{adjustbox}{max width=0.9\linewidth}
      \begin{tabular}{ccc|ccc}
        \toprule
        \multicolumn{3}{c|}{PAM} \\
        layer4 & layer5 & layer6 & $mAP_{25}$ & $mAP_{50}$ & $mAP_{75}$ \\
        \hline
                     &              &              &  40.2  &  34.7   &  19.6  \\
                     &              & \checkmark   &  \textbf{53.5}  &  \textbf{41.8}   &  \textbf{24.2}  \\
                     &  \checkmark  & \checkmark   &  49.6  &  39.5   &  23.9  \\
         \checkmark  &  \checkmark  & \checkmark   &  43.7  &  34.8   &  21.2  \\
         \checkmark  &              & \checkmark   &  53.6  &  40.0   &  23.8  \\
         \checkmark  &              &              &  48.3  &  37.6   &  22.1  \\
                     &  \checkmark  &              &  49.9  &  39.6   &  23.4  \\
         \checkmark  &  \checkmark  &              &  49.3  &  38.4   &  22.7  \\
         \bottomrule
      \end{tabular}
    \end{adjustbox}
    \label{tab:pam_module}
\end{table}

\section*{Appendix: Qualitative Results of Pseudo Label}
\begin{figure*}[t]
    \centering
    \includegraphics[width=\linewidth]{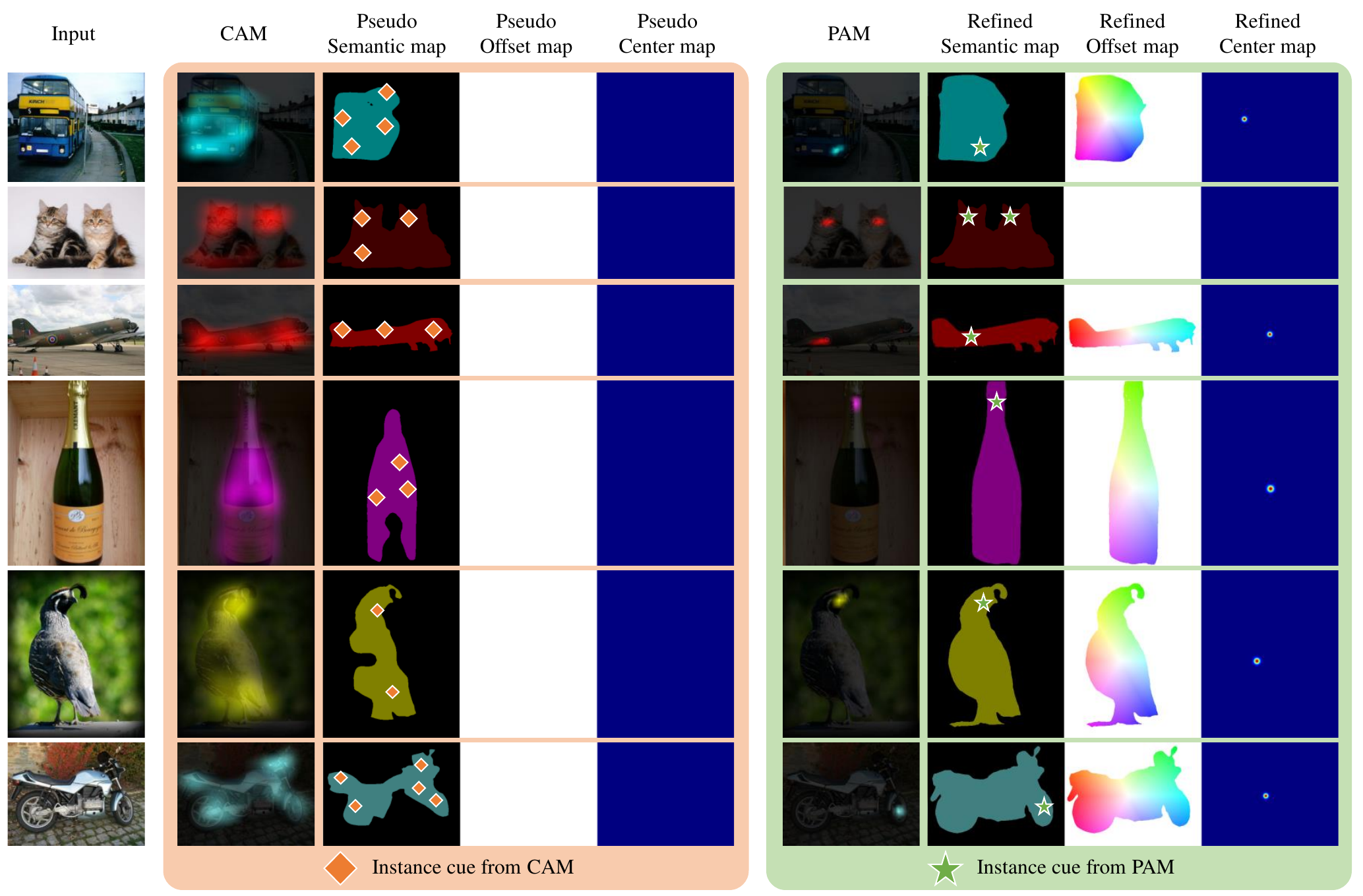}
    \caption{Qualitative results of pseudo labels generated from conventional CAMs (orange region) and pseudo labels generated from our PAM (green region). The PAM can more accurately extract one peak point per instance than the CAM.}
    \label{fig:cam_and_label}
\end{figure*}

In Figure \ref{fig:cam_and_label}, we provide more qualitative results of activation maps and pseudo labels.
As in the orange area of Figure \ref{fig:cam_and_label}, the conventional CAMs have a limitation in generating high-quality pseudo labels due to the noisy activation region.
However, as in the green area of Figure \ref{fig:cam_and_label}, our PAM produces sparse CAMs that help to extract one instance cue per instance.
Therefore, from the \textit{semantic knowledge transfer}, we can obtain more reliable pseudo labels, and the pseudo labels contain more true positive training samples.

\section*{Appendix: Qualitative Results of Proposed Method}
\begin{figure*}[t]
    \centering
    \includegraphics[width=\linewidth]{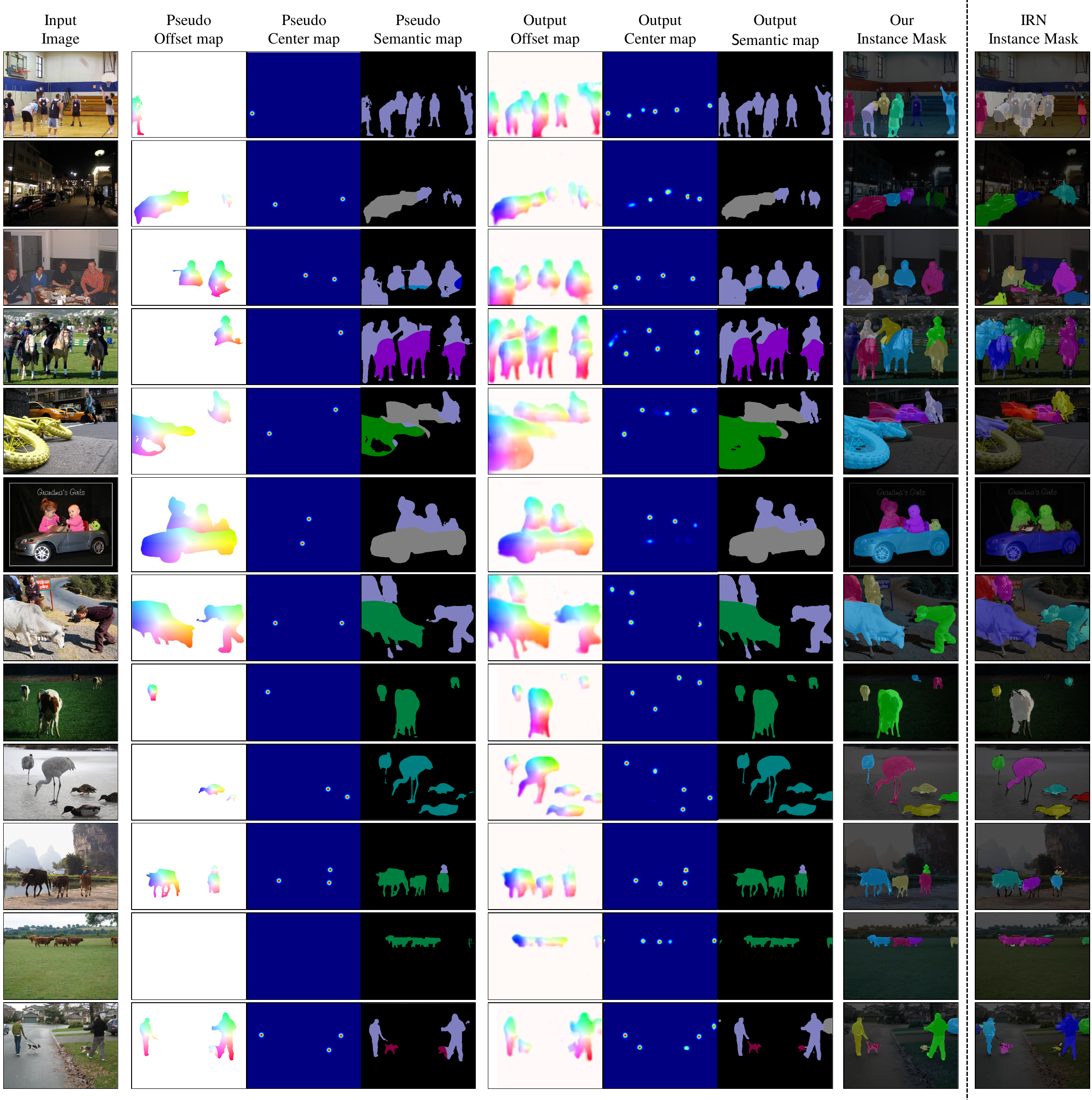}
    \caption{Qualitative results of our pseudo labels and outputs of BESTIE on VOC 2012 dataset. We note that we only use the image-level labels without the off-the-shelf proposal techniques. 
    Compared with IRN~\cite{(irn)ahn2019weakly}, which is the proposal-free method, our BESTIE can segment multiple instances more accurately and precisely.}
    \label{fig:label_and_mask}
\end{figure*}

In Figure \ref{fig:label_and_mask}, we provide more qualitative results of our pseudo labels and network outputs.
The pseudo label provides some reliable true-positive samples but contains lots of false-negatives ($i.e.,$ missing instances).
Due to the proposed \textit{self-refinement} with the \textit{instance-aware guidance}, the network can produce high-quality instance masks including missing instances in pseudo labels.
In addition, we compare our instance mask with that of IRN~\cite{(irn)ahn2019weakly}, which is the proposal-free method.
The comparison results clearly show that our approach can properly segment multiple instances with a high-precision instance mask.

\section*{Appendix: Failure Cases for PAM}
\begin{figure}[t]
    \centering
    \includegraphics[width=\linewidth]{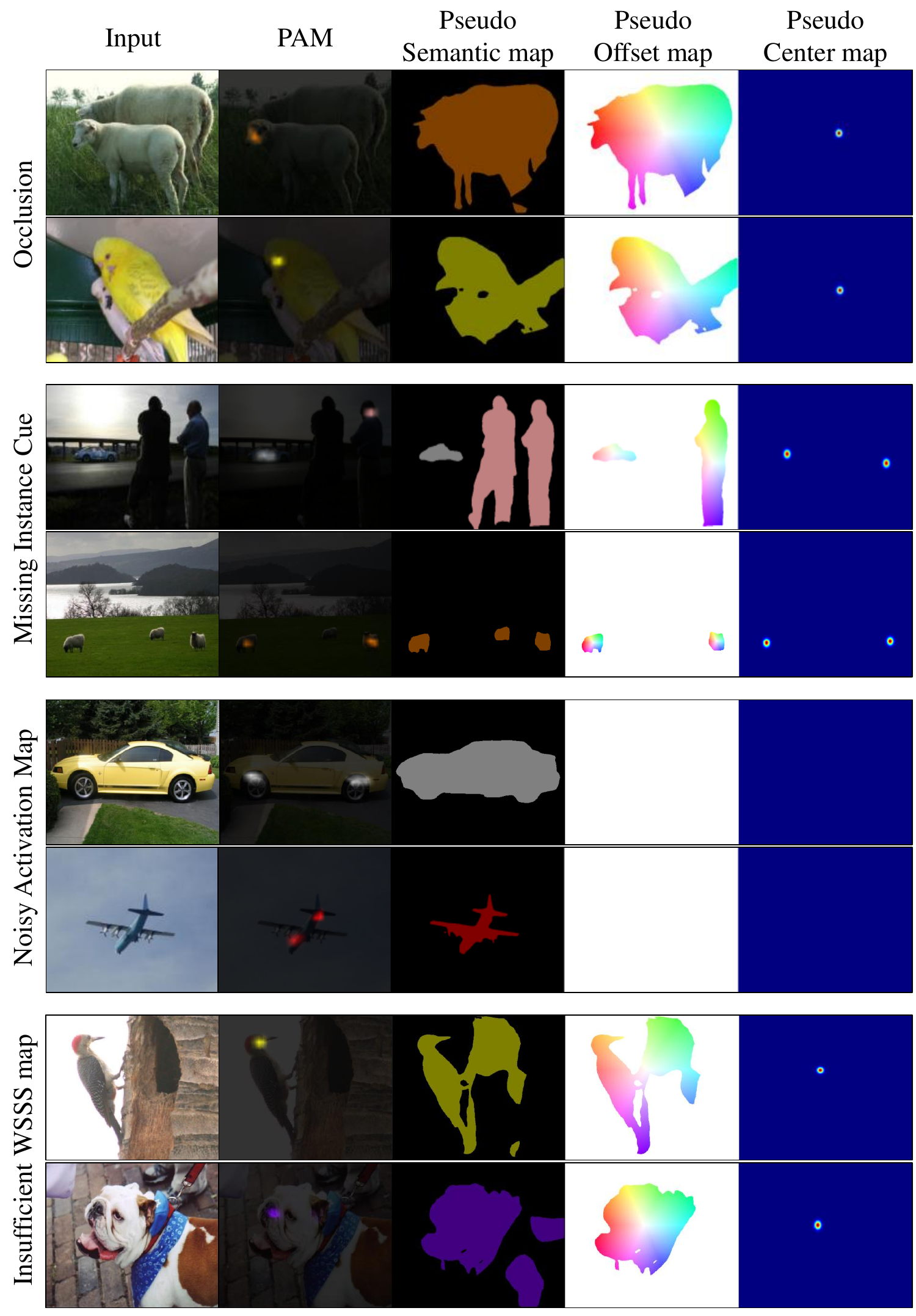}
    \caption{Failure cases of the PAM.}
    \label{fig:failure_PAM}
\end{figure}

We provide some failure cases of PAM in Figure \ref{fig:failure_PAM}, and these examples demonstrate the superiority of the point-supervised setting because inaccurate instance cues are replaced by ground-truth points.
PAM has trouble in accurately localizing overlapping instances, which leads to the incorrect pseudo label (first row in Figure \ref{fig:failure_PAM}).
In addition, missing instance cue or noisy localization increases missing instances in the pseudo label (second and third rows in Figure \ref{fig:failure_PAM}).
Last, the WSSS method is trained with only image-level labels, some insufficient semantic segmentation maps yield inaccurate pseudo labels (last row in Figure \ref{fig:failure_PAM}).

\section*{Appendix: Failure Cases for Proposed Method}
\begin{figure}[t]
    \centering
    \includegraphics[width=\linewidth]{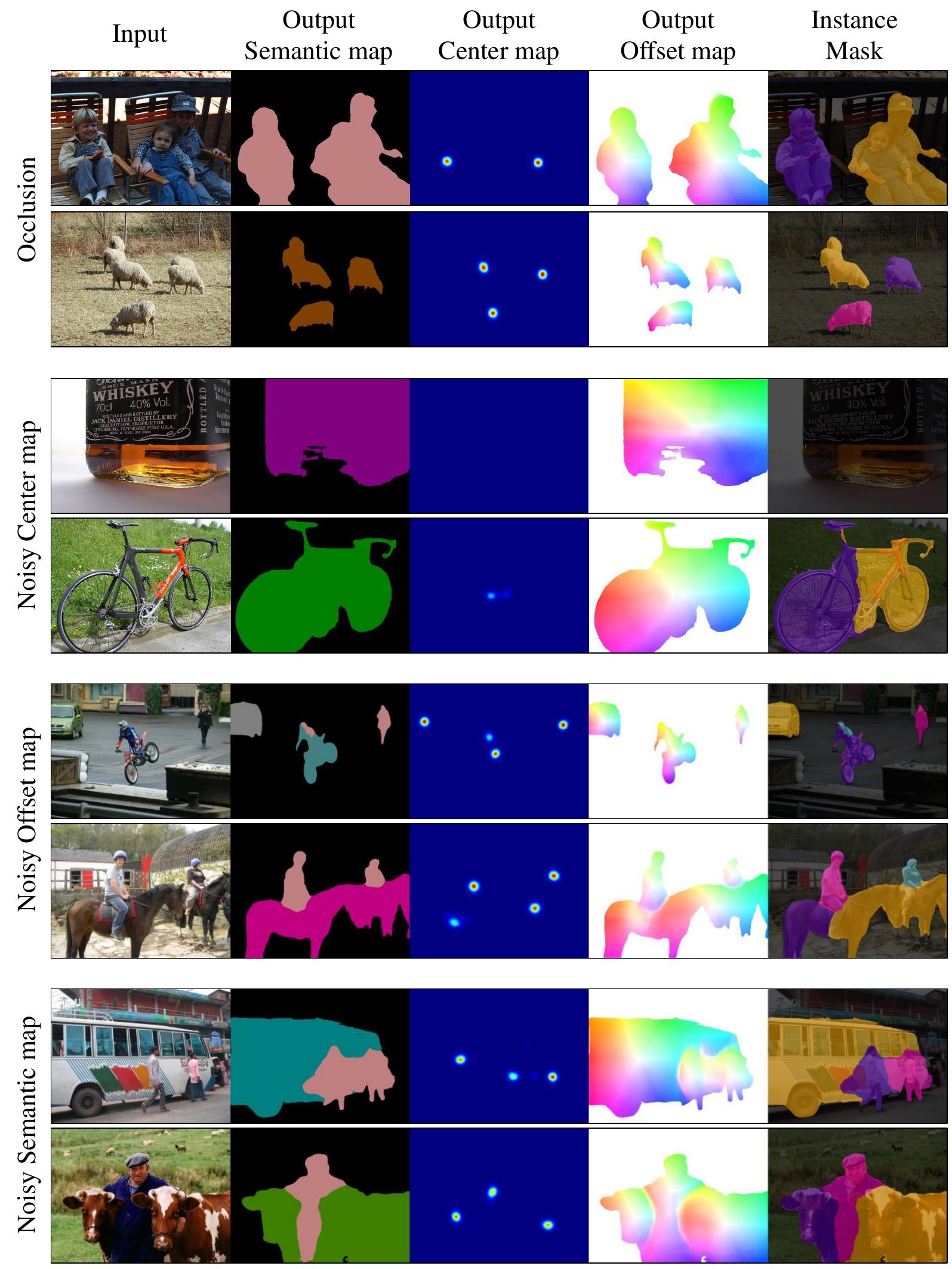}
    \caption{Failure cases of BESTIE.}
    \label{fig:failure_Inst}
\end{figure}

We provide some failure cases of BESTIE in Figure \ref{fig:failure_Inst}.
First, when center points of instances are close to each other, we often fail to obtain the proper instance masks (first row in Figure \ref{fig:failure_Inst}); however, the keypoint-based method has suffered this issue even in a fully-supervised setting.
In addition, noisy center and offset maps lead to false instance masks (second and third rows in Figure \ref{fig:failure_Inst}).
Last, when the semantic segmentation map provides a noisy foreground region, we often fail to obtain the precise instance mask.

\section*{Appendix: Pytorch-style Pseudo-code.}
To describe the details of each proposed methods, we provide pytorch-style pseudo-code algorithm.
Note that our BESTIE is simple and easy to be implemented.

\begin{lstlisting}[language=Python, caption=Overview of BESTIE]
    for n in range(train_iterations):
        # load data: x (input tensor), seg_map (weakly-supervised semantic map)
        x, seg_map = loader.next()
    
        # pseudo-label generation
        PAM = classifier(x)
        peak_points = extract_peak_points(PAM)
        pseudo_label = pseudo_label_gen(seg_map, peak_points)
    
        # network forwarding
        outputs = BESTIE(x)
    
        # refined-label generation
        refined_label = refined_label_gen(outputs)
    
        # backward & optimize
        loss_pseudo = objective_function_with_IAG(outputs, pseudo_label)
        loss_refined = objective_function_with_IAG(outputs, refined_label)
        loss = loss_pseudo + loss_refined
        loss.backward()
        optimizer.step()
\end{lstlisting}

\begin{lstlisting}[language=Python, caption=PAM Module]
    def PAM(x, control_values):
        x = F.relu(x) # [B, K, H, W]
        
        criteria_points = F.adaptive_max_pool2d(x, 1) # [B, K, 1, 1]
        criteria_points = criteria_points.expand_as(x) # [B, K, H, W]
        
        noisy_region = (x < criteria_points * control_values)
        x[noisy_region] = 0 # deactivate noisy region
        return x
\end{lstlisting}

\begin{lstlisting}[language=Python, caption=Point Extraction]
    def extract_peak_points(heatmap, kernel, threshold, K=40):
        B, C, H, W = heatmap.size()
        
        heatmap_max = F.max_pool2d(heatmap, (kernel, kernel), 
                                  stride=1, padding=(kernel - 1) // 2)
        keep = (heatmap_max == heatmap).float()
        
        local_max = heatmap * keep
        peak_points = torch.nonzero(local_max > threshold)
        return peak_points
\end{lstlisting}

\begin{lstlisting}[language=Python, caption=Pseudo-Label Generation]
    def pseudo_label_gen(seg_map, peak_points):
        # seg_map : [H, W] , peak points : [N, 2]
        mask_candidates = connected_component_labeling(seg_map)
        
        for mask in mask_candidates:
            if one_peak_point_in_mask(mask, peak_points):
                center_point = centroid(mask)
                center_map_generation(center_point)
                offset_map_generation(center_point, mask)
    
    
    def offset_map_generation(center_point, mask): # mask: bindary mask
        cy, cx = center_point
        mask_idx = np.where(mask > 0)
        
        coord = np.ones_like(mask, dtype=np.float32)
        y_coord = np.cumsum(coord, axis=0)-1
        x_coord = np.cumsum(coord, axis=1)-1
        
        offset_y_index = (np.zeros_like(mask_idx[0]), mask_idx[0], mask_idx[1])
        offset_x_index = (np.ones_like(mask_idx[0]), mask_idx[0], mask_idx[1])
        
        pseudo_offset_map[offset_y_index] = cy - y_coord[mask_idx]
        pseudo_offset_map[offset_x_index] = cx - x_coord[mask_idx]
\end{lstlisting}

\begin{lstlisting}[language=Python, caption=Refined Label Generation]
    def refined_label_gen(center_map, offset_map, seg_map, thresh):
        instance_masks = instance_grouping(center_map, offset_map, seg_map, thresh)
    
        for mask in instance_masks:
            center_point = centroid(mask)
            center_map_generation(center_point)
            offset_map_generation(center_point, mask)
            
            
    def instance_grouping(center_map, offset_map, seg_map, thresh):
        ctr = extract_peak_points(center_map, thresh)
        ctr = center_clustering(ctr, offset_map)
        # center_map : [1, C, H, W], ctr : [N, 2], offset_map : [2, H, W]
    
        H, W = offset_map.size()[1:]
        y_coord = torch.arange(H).repeat(1, W, 1).transpose(1, 2)
        x_coord = torch.arange(W).repeat(1, H, 1)
        coord = torch.cat((y_coord, x_coord), dim=0)
    
        ctr_loc = coord + offsets
        ctr_loc = ctr_loc.reshape((2, H*W)).transpose(1, 0) # [H*W, 2]
    
        dist = torch.norm(ctr.unsqueeze(1)-ctr_loc.unsqueeze(0), dim=-1) # [N, H*W]
    
        # finds center with minimum distance at each location
        instance_mask = torch.argmin(dist, dim=0).reshape((1, H, W)) + 1
    
        return instance_mask
\end{lstlisting}

\begin{lstlisting}[language=Python, caption=Objective Function with IAG]
    def objective_function_with_IAG(pred, gt, eps=1e-4):
        gamma_offset, gamma_center, gamma_semantic = 0.01, 200.0, 1.0
        guidance_region = (gt['offset'] != 0).float() # labeled instance region
    
        offset_map_loss = F.l1_loss(pred['offset'], gt['offset'], reduction='none') * guidance_region
        offset_map_loss = offset_map_loss.sum() / (guidance_region.sum() + eps)
    
        center_map_loss = F.mse_loss(pred['center'],gt['center'], reduction='none') * guidance_region
        center_map_loss = center_map_loss.sum() / (guidance_region.sum() + eps)
    
        semantic_map_loss = F.cross_entropy(pred['semantic'], gt['semantic'])
    
        return (offset_map_loss * gamma_offset) + (center_map_loss * gamma_center) + (semantic_map_loss * gamma_semantic)
\end{lstlisting}

\end{document}